\documentclass[10pt,twocolumn,letterpaper]{article}

\usepackage{cvpr}
\usepackage{times}
\usepackage{epsfig}
\usepackage{graphicx}
\usepackage{amsmath}
\usepackage{amssymb}
\usepackage{multirow}
\usepackage{booktabs} 
\usepackage{algorithm}
\usepackage{algorithmic}
\usepackage{subfigure}

\usepackage{dsfont}

\usepackage[pagebackref=true,breaklinks=true,letterpaper=true,colorlinks,bookmarks=false]{hyperref}
\cvprfinalcopy 

\def\cvprPaperID{347} 

\ifcvprfinal\fi
\begin{document}

\title{Interpretable Structure-Evolving LSTM}

\author{Xiaodan~Liang$^{1}$ \quad Liang~Lin$^{2}$ \quad Xiaohui Shen$^{4}$ \quad Jiashi Feng$^{3}$  \quad Shuicheng~Yan $^{3}$ \quad Eric P. Xing$^{1}$\\
$^{1}$ Carnegie Mellon University \quad $^{2}$ Sun Yat-sen University \quad $^{3}$ National University of Singapore\\ \quad  $^{4}$ Adobe Research\\
\scriptsize{xiaodan1@cs.cmu.edu, linliang@ieee.org, xshen@adobe.com, elefjia@nus.edu.sg,  eleyans@nus.edu.sg, epxing@cs.cmu.edu}
}

\maketitle

\begin{abstract}
	
This paper develops a general framework for learning interpretable data representation via Long Short-Term Memory (LSTM) recurrent neural networks over hierarchal graph structures. Instead of learning LSTM models over the pre-fixed structures, we propose to further learn the intermediate interpretable multi-level graph structures in a progressive and stochastic way from data during the LSTM network optimization. We thus call this model the structure-evolving LSTM. In particular, starting with an initial element-level graph representation where each node is a small data element, the structure-evolving LSTM gradually evolves the multi-level graph representations by stochastically merging the graph nodes with high compatibilities along the stacked LSTM layers. In each LSTM layer, we estimate the compatibility of two connected nodes from their corresponding LSTM gate outputs, which is used to generate a merging probability. The candidate graph structures are accordingly generated where the nodes are grouped into cliques with their merging probabilities. We then produce the new graph structure with a Metropolis-Hasting algorithm, which alleviates the risk of getting stuck in local optimums by stochastic sampling with an acceptance probability. Once a graph structure is accepted, a higher-level graph is then constructed by taking the partitioned cliques as its nodes. During the evolving process, representation becomes more abstracted in higher-levels where redundant information is filtered out, allowing more efficient propagation of long-range data dependencies. We evaluate the effectiveness of structure-evolving LSTM in the application of semantic object parsing and demonstrate its advantage over state-of-the-art LSTM models on standard benchmarks.
	
\end{abstract}

\section{Introduction}

Recently, there has been a surge of interest in developing various kinds of Long Short-Term Memory (LSTM) neural networks for modeling complex dependencies within sequential and multi-dimensional data, due to their advantage in a wide range of applications such as speech recognition~\cite{graves2013speech}, image generation~\cite{van2016pixel}, image-to-caption generation~\cite{showtell} and multi-dimensional image processing~\cite{gridlstm}. 

  \begin{figure*}[!tp]
  	\begin{center}
  		\includegraphics[scale=0.6]{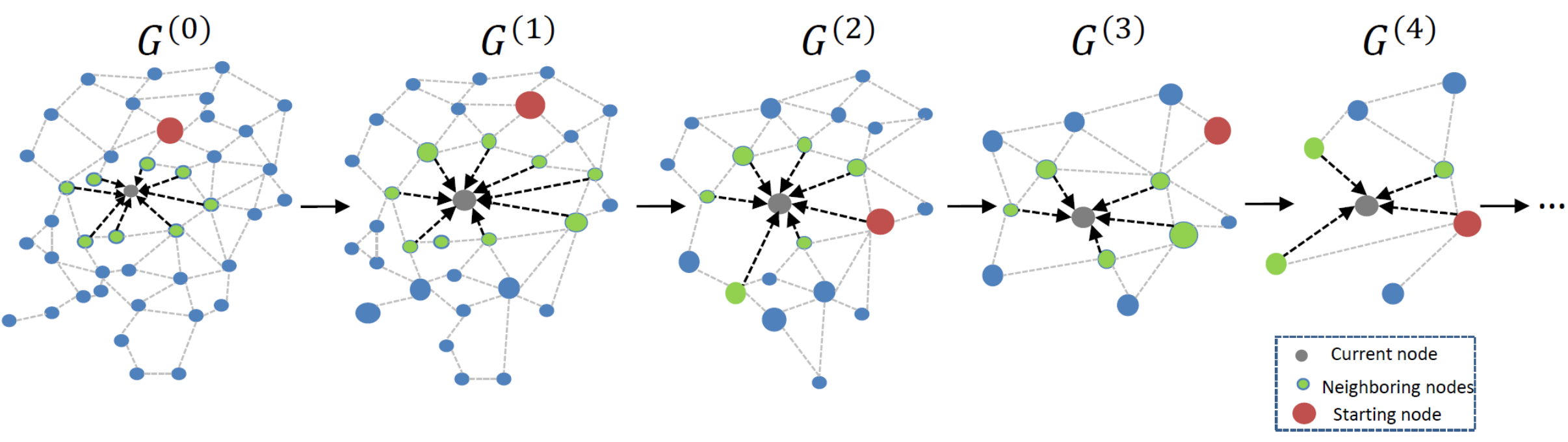}
  		\caption{{An illustration of the structure evolving process of the proposed structure-evolving LSTM model. Starting from an initial graph $G^{(0)}$, the structure-evolving LSTM learns to evolve the hierarchical graph structures with a stochastic and bottom-up node merging process and then propagates the information on these generated multi-level graph topologies following a stochastic node updating scheme.}}
  		\label{fig:graphlstm}
  		\vspace{-6mm}
  	\end{center}
  \end{figure*}
  
Despite the remarkable success, existing LSTM models such as chain-structured~\cite{graves2013speech}~\cite{showtell}, tree-structured LSTM models~\cite{zhu2015long,tai2015improved} and graph-structured LSTM~\cite{liang2016semantic} can only process data with pre-fixed structures in terms of their internal information propagation route. They are therefore limited in dealing with the data containing complex multi-level correlations. For example, the structure of human social network is inherently hierarchical, where each individual is a member of several communities, ranging from small (e.g., families, friends) to large (e.g., organizations such as schools and businesses). Semantic object parsing in an image, for another example, can benefit from modeling the contextual dependencies among regions in different levels, where the lower-level graph representation on small regions (e.g., superpixels) can preserve the local and fine object boundaries while the higher-level graph on larger coherent regions captures more semantic interactions. Thus, in order to well abstract multi-level representations of such data, it is desirable to integrate the data structure evolving with LSTM parameter learning. 

In this work, we seek a general and interpretable framework for representing the data via LSTM networks over the dynamically learned multi-level data structures, in which hierarchical intrinsic representations are simultaneously learned from the data along with encoding the long-term dependencies via LSTM units. Since numerous important problems can be framed as learning from graph data (tree-structure can be treated as one specific graph), our structure-evolving directly investigates the hierarchical representation learning over the initial arbitrary graph structures. However, learning dynamic hierarchical graphs is much more challenging than the convenient hierarchical convolution neural networks due to the arbitrary number of nodes, orderless node layouts and diverse probabilistic graph edges.  To learn intermediate interpretable graph structures of the data and alleviate the over-fitting problem, we design a stochastic algorithm to sample the graph structure (i.e., the grouping of graph nodes) in each LSTM layer and gradually build the multi-level graph representations in a bottom-up manner. We thus name our model as the structure-evolving LSTM. Compared with existing LSTM structures with pre-fixed chain~\cite{graves2013speech}~\cite{showtell}, tree~\cite{zhu2015long,tai2015improved} or graph topologies~\cite{liang2016semantic}, the structure-evolving LSTM has the capability of modeling long-range interactions using the dynamically evolved hierarchical graph topologies to capture the multi-level inherent correlations embedded in the data.

As illustrated in Fig.~\ref{fig:graphlstm}, the structure-evolving LSTM gradually evolves the multi-level graph representations through a stochastic and bottom-up node merging process, starting with an initial graph in which each node indicates a data element and every two neighboring nodes are linked by an edge. To enable learn the interpretable hierarchical representation, we propose to progressively merge different graph nodes guided by the global advantage reward at each step. The new graph that is composed by the merged graph nodes and updated graph edges is thus generated by a stochastic policy that ensures not only the less overhead graph transition from the previous graph to the new graph and the advantage discriminative capability brought by the new graph. 

Specifically, for two connected nodes, their merging probability is estimated from the adaptive forget gate outputs in the LSTM unit, indicating how likely the two nodes tend to be merged into a clique (i.e., a node at the higher level graph). Then the graph structure is generated by designing a Metropolis-Hasting algorithm~\cite{barbu2003graph,tu2002image}. Specifically, this algorithm stochastically merging some graph nodes by sampling their merging probabilities, and produces a new graph structure (i.e., a set of partitioned cliques). This structure is further examined and determined according to a global reward defined as an acceptance probability. Under such a stochastic sampling paradigm, the acceptance probability involves two terms: i) a state transition probability (i.e., a product of the merging probabilities); ii) a posterior probability representing the compatibility of the generated graph structure with task-specific observations. Intuitively, this global reward thus encourages the structure-evolving step that better not leads to a hugh graph shift (i.e., only very few edges are merges) and also can help boost the target-specific performance.

Once a new level of graph structure is evolved, the LSTM layer broadcasts information along the generated graph topology following a stochastic updating scheme, in order to enable global reasoning on all nodes. In turn, the updated LSTM gate outputs induce the merging probability of graph nodes for the subsequent graph structure evolving. Instead of being influenced equally by all of its neighboring nodes in each LSTM unit, our model learns the adaptive forget gates for each neighboring node when updating the hidden states of a certain node. Such an adaptive scheme has advantage in conveying semantically meaningful interactions between two graph nodes. The network parameters are then updated by back-propagation in an end-to-end way.

We leverage the structure-evolving LSTM model to address the fundamental semantic object parsing task and experimentally show that structure-evolving LSTM outperforms other state-of-the-art LSTM structures on three object parsing datasets.

 \begin{figure*}[!tp]
 	\begin{center}
 		\includegraphics[scale=0.7]{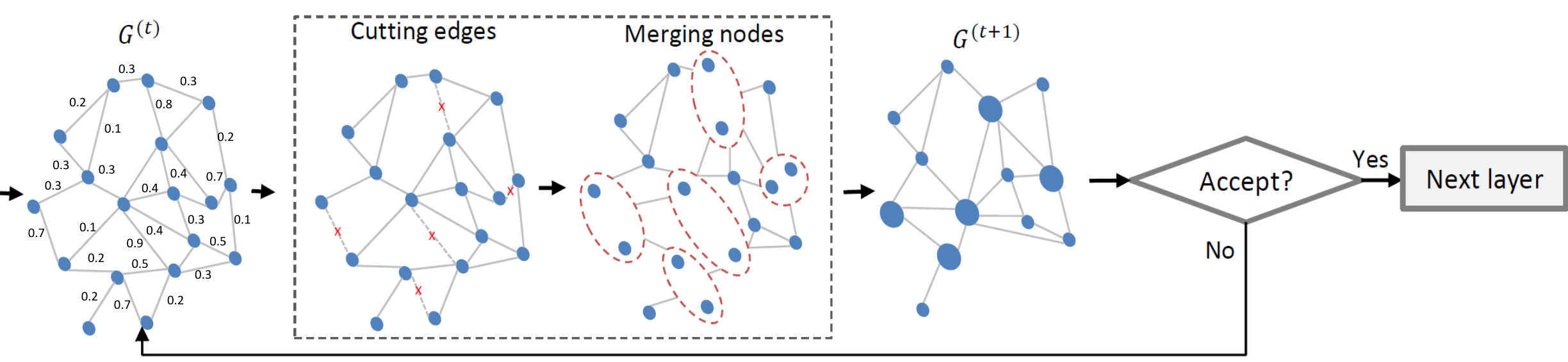}
 		\caption{{Illustration of the stochastic structure-evolving step for evolving a lower-level graph into a higher-level one. Given the computed merging probabilities for all nodes, our structure-evolving step takes several trials to evolve a new graph till the new graph is accepted evaluated by the acceptance probability. A new graph is generated by stochastically merging two nodes with high predicted merging probabilities and thus the new edges are produced. The acceptance probabilities are computed by considering the graph transition cost and the advantage discriminative capability brought by the new graph.}}
 		\label{fig:evolving}
 		\vspace{-6mm}
 	\end{center}
 \end{figure*}

\section{Related Works}

Long Short-Term Memory (LSTM) recurrent networks have been first introduced to address the sequential prediction tasks~\cite{graves2009offline,sutskever2014sequence,showtell,lstmempirical}, and then extended to multi-dimensional image processing tasks~\cite{byeon2014texture,theis2015generative} such as image generation~\cite{van2016pixel,theis2015generative}, person detection~\cite{stewart2015end}, scene labeling~\cite{byeon2015scene} and object parsing~\cite{liang2015semantic}. It can keep long-term
memory by training proper gating weights and has practically showed the effectiveness on a range of problems~\cite{byeon2014texture,byeon2015scene}. 
For image processing, in each LSTM unit, the prediction of each pixel is designed to affected by a fixed factorization (e.g., 2 or 8 neighboring pixels~\cite{gridlstm}\cite{MDLSTM}\cite{liang2015semantic} or diagonal neighborhood~\cite{van2016pixel}\cite{theis2015generative}). Recently, Tree-LSTM~\cite{tai2015improved} introduces the structure with tree-structured topologies for predicting semantic representations of sentences. Graph LSTM~\cite{liang2016semantic} has been proposed to propagate information on a basic pre-defined graph topology to capture diverse natural visual correlations (e.g., local boundaries and homogeneous regions). However, the complex patterns in different modalities often embed hierarchal structures, representing different levels of correlations between nodes. Different from using pre-defined data structures in previous LSTMs~\cite{gridlstm,MDLSTM,liang2015semantic,liang2016semantic,theis2015generative}, the proposed structure-evolving LSTM targets on automatically learning the hierarchical graph representations by evolving from an initial graph structure. In this way, the intrinsic multi-level semantic abstractions can be learned and then used to boost the multi-scale reasoning by LSTM units.   

The structure-evolving LSTM (dynamically evolving multi-level graphs) is superior to the most related Graph LSTM~\cite{liang2016semantic} (a pre-fixed single-level graph) in two aspects: 1) Structure-evolving LSTM learns more powerful representations as it progressively exploits hierarchical information along stacked LSTM layers; 2) at its later layers, the structure-evolving LSTM captures the inherent structure of the desired output benefiting from the higher-level graph topologies. These superiorities bring significant improvements on several semantic parsing datasets, which gives apple-to-apple comparison with~\cite{liang2016semantic}. Our work aims to develop a new and general principled graph evolving based learning method to learn more powerful Graph LSTM or other RNN models. Devising new Graph-LSTM unit is not within the scope of this paper. We use Graph-LSTM as a running example which by no means implies our method is limited to Graph LSTM.

\section{Structure-evolving LSTM}

Fig.~\ref{fig:graphlstm} illustrates the proposed structure-evolving LSTM network architecture. Suppose that the initialized graph for the data is denoted as $G^{(0)} =<V ^{(0)}, \mathcal{E}^{(0)} >$, where $V ^{(0)}$ and $\mathcal{E}^{(0)}$ are the corresponding graph nodes (e.g., data elements) and edges. Each node $v^0_i \in V^{(0)}, \{i \in 1, \cdots, N^0\}$ is represented by the deep features $f^{(0)}_i$ learned from the underlying CNN model with D dimensions. Based on the LSTM gate outputs and the graph $G^{(t)}$ in the previous $t$-th LSTM layer, structure-evolving LSTM then learns a higher-level graph structure $G^{(t+1)} =< V^{(t+1)}, \mathcal{E}^{(t+1)} > $ for the information propagation in the $(t + 1)$-th LSTM layer. Learning new graph structures and updating LSTM parameters are thus alternatively performed and the network parameters are trained in an end-to-end way.

\subsection{Basic LSTM Units}

Given the dynamically constructed graph structure $G^{(t)}$, the $t$-th structure-evolving LSTM layer determines the states of each node $v^t_i$ that comprise the hidden states $\mathbf{h}^t_i$  and the memory states $\mathbf{m}^t_i$ of each node, and the edge probability $p^t_{ij}$ of two nodes for evolving a new graph structure. The state of each node is influenced by its previous states and the states of connected graph nodes in order to propagate information to all nodes. Thus the inputs to LSTM units consist of the input states $\mathbf{f}^t_i$ of the node $v^t_i$ , its previous hidden states $\mathbf{h}^{(t-1)}_i$ and memory states $\mathbf{m}^{(t-1)}_i$, and the hidden and memory states of its neighboring nodes $v^t_j, j\in \mathcal{N}_{G^{(t)}}(i)$. Note that there is a flexibility in the order of node updating in the structure-evolving LSTM layers. Following ~\cite{liang2016semantic}, we randomly specify the node updating sequence to propagate information to all nodes in order to increase the model diversity during learning the LSTM network parameters. 

Our structure-evolving LSTM follows the Graph LSTM units~\cite{liang2016semantic} to generate hidden and memory cells, and then show how to inject the edge merging probabilities of the nodes into the LSTM units. We thus first introduce the generation of hidden and memory cells to make this paper more self-contained. When operating on a specific node $v^t_i$, some of its neighboring nodes have already been updated while others may not. We therefore use a visit flag $q^t_j$ to indicate whether the graph node $v^t_j$ has been updated, where $q^t_j$ is set as 1 if updated and otherwise 0. We then use the updated hidden states $\mathbf{h}^t_j$ for the visited nodes and the previous states $\mathbf{h}^{t-1}_j$ for the unvisited nodes. Note that the nodes in the graph may have an arbitrary number of neighboring nodes. Let $\mathcal{N}_{G^{(t)}}(i)$ denote the number of neighboring graph nodes for the node $i$. To obtain a fixed feature dimension for the inputs of the Graph LSTM unit during network training, the hidden states $\mathbf{\bar{h}}^{t-1}_i$ used for computing the LSTM gates of the node $v^t_i$ are obtained by averaging the hidden states of neighboring nodes, computed as:
{\small
\begin{equation}
\bar{\mathbf{h}}_{i}^{t-1} = \frac{\sum_{j \in \mathcal{N}_{{G^{(t)}}}(i) }(\mathds{1}(q_j^t = 1)\mathbf{h}_{j}^{t} + \mathds{1}(q_j^t = 0)\mathbf{h}_{j}^{t-1})}{|\mathcal{N}_{{G^{(t)}}}(i)|}.
\end{equation}}
	
\textbf{Structure-evolving LSTM.} The structure-evolving LSTM consists of five gates:  the input gate $g^u$, the forget gate $g^f$, the adaptive forget gate $\bar{g}^f$, the memory gate $g^c$, the output gate $g^o$ and the edge gate $p$. The $\mathds{1}$ is an indicator function. The $W^e$ indicates the recurrent edge gate weight parameters. The $W^u, W^f, W^c, W^o$ are the recurrent gate weight matrices specified for input features while $U^u,U^f, U^c, U^o$ are those for hidden states of each node. $U^{un}, U^{fn}, U^{cn}, U^{on}$ are the weight parameters specified for states of neighboring nodes. The structure-evolving LSTM unit specifies different forget gates for different neighboring nodes by functioning the input states of the current node with their hidden states, defined as $\bar{g}^f_{ij}, j \in \mathcal{N}_{{G^{(t)}}}(i)$. It results in the different influences of neighboring nodes on the updated memory states $\mathbf{m}_{i}^{t+1}$ and hidden states $\mathbf{h}_{i}^{t+1}$. The merging probability $p_{ij}$ of each pair of graph nodes $<i,j> \in \mathcal{E}^{(t)}$ is calculated by weighting the adaptive forget gates $\bar{g}^f_{ij}$ with the weight matrix $W^e$. Intuitively, adaptive forget gates are to identify the distinguished correlations of different node pairs, e.g. some nodes have stronger correlations than others. The merging probability for each pair is thus estimated from adaptive forget gates for graph evolving. The new hidden states, memory states and edge gates (i.e., merging probabilities of each connected pair of nodes) in the graph $G^{(t)}$ can be calculated as follows:

{\small
\begin{equation}
	\begin{split}
	g^u_i = &\delta(W^u\mathbf{f}_{i}^{t} + U^u\mathbf{h}_{i}^{t-1} + U^{un}\bar{\mathbf{h}}_{i}^{t-1} + b^u),\\
	\bar{g}^f_{ij} =& \delta(W^f\mathbf{f}_{i}^{t} + U^{fn}\mathbf{h}_{j}^{t-1} + b^f),\\
	g^f_{i} = &\delta(W^f\mathbf{f}_{i}^{t} + U^f\mathbf{h}_{i}^{t-1} + b^f),\\
	g^o_i = &\delta(W^o\mathbf{f}_{i}^{t} + U^o\mathbf{h}_{i}^{t-1} + U^{on}\bar{\mathbf{h}}_{i}^{t-1} + b^o),\\
	g^c_i = &\tanh(W^c\mathbf{f}_{i}^{t}  + U^c\mathbf{h}_{i}^{t-1} + U^{cn}\bar{\mathbf{h}}_{i}^{t-1} + b^c),\\
	\mathbf{m}_{i, t} = &\frac{\sum_{j \in \mathcal{N}_{\mathcal{G}}(i)}(\mathds{1}(q_j = 1) \bar{g}^f_{ij} \odot \mathbf{\mathbf{m}}_{j}^{t} + \mathds{1}(q_j = 0)\bar{g}^f_{ij} \odot \mathbf{\mathbf{m}}_{j}^{t-1})}{|\mathcal{N}_{{G^{(t)}}}(i)|}\\ 
	& +  g^f_{i}\odot \mathbf{\mathbf{m}}_{i}^{t-1} +  g^u_i \odot g^c_i,\\
	\mathbf{h}_{i}^{t} =& \tanh(g^o_i \odot \mathbf{m}_{i}^{t})\\
	p_{ij}^t = &\delta(W^e \bar{g}^f_{ij}).
	\end{split}
	\label{eq:lstm}
\end{equation}}

\noindent{Here} $\delta$ is a logistic sigmoid function, and $\odot$ indicates a point-wise product. Let $\mathbf{W}, \mathbf{U}$ denote the concatenation of all weight matrices and $\{\mathbf{Z}_{j,t}\}_{j \in \mathcal{N}_{\mathcal{G}}(i)}$ represent all the related information of neighboring nodes. This mechanism acts as a memory system, where the information can be written into the memory states and sequentially recorded by each graph node, which is then used to communicate with the hidden states of subsequent graph nodes and previous LSTM layer. And the merging probabilities $\{p_{ij}\}, <i,j> \in \mathcal{E}^{(t)}$ can be conveniently learned and used for generating the new higher-level graph structure $G^{(t+1)}$ in the $(t+1)$-th layer, detailed in Section~\ref{sec:evolv}. During training, the merging probabilities of graph edges are supervised by approximating to the final graph structure for a specific task, such as the connections of final semantic regions for image parsing. The back propagation is used to train all the weight metrics.

\subsection{Interpretable Structure Evolving}
\label{sec:evolv}

Given the graph structure $G^{(t)} = <V^{(t)}, \mathcal{E}^{(t)}>$ and all merging probabilities  $\{p_{ij}\}, <i,j> \in \mathcal{E}^{(t)}$, the higher-level graph structure $G^{(t+1)}$ can be evolved by stochastically merging some graph nodes and examined with an acceptance probability, as shown in Fig.~\ref{fig:evolving}. Specifically, a new graph node $G^{(t+1)}$ is constructed by merging some graph nodes with the merging probability. As there is no deterministic graph transition path from an initial graph to the final one, it is intractable to enumerate all possible $G^{(t+1)}$ for evaluation within the large search space. We thus use a stochastic
mechanism rather than a deterministic one to find a good graph transition. Such a stochastic searching scheme is also effective in alleviating the risk of getting trapped in a bad local optimum. To find a better graph transition between two graphs $G^{(t)}$ and $G^{(t+1)}$, the acceptance rate of the transition from the graph  from $G^{(t)}$ to graph $G^{(t+1)}$ is defined by a Metropolis-Hastings method~\cite{barbu2003graph,tu2002image}:

\begin{equation}
		\begin{split}
	\alpha(G^{(t)}\rightarrow G^{(t+1)}) = &\min(1, \\\frac{q(G^{(t+1)\rightarrow G^{(t)}})}{q(G^{(t)\rightarrow G^{(t+1)}})}
	&\frac{P(G^{(t+1)}|I; \mathbf{W}, \mathbf{U})}{P(G^{(t)}|I; \mathbf{W}, \mathbf{U})}).
		\end{split}
	\end{equation}

\begin{figure*}[!tp]
	\begin{center}
		\includegraphics[scale=0.65]{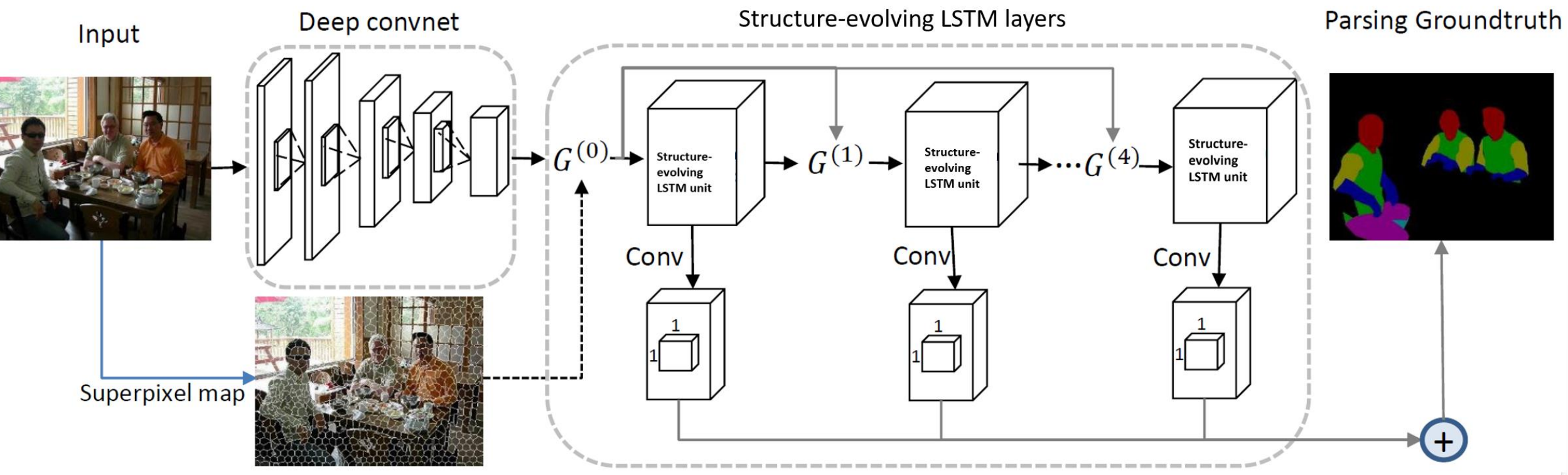}
		\caption{{Overview of the segmentation network architecture that employs the structure-evolving LSTM layer for semantic  object parsing in image domain. Based on the basic convolutional feature maps, five structure-evolving LSTM layers are stacked to propagate information on the stochastically generated multi-level graph structures (i.e., $G^{(0)}, G^{(1)}, G^{(2)}, G^{(3)}, G^{(4)}$) where $G^{(0)}$ is constructed as the superpixel neighborhood graph. The convolutional layers are appended on all LSTM layers to produce the multi-scale predictions, which are then combined to generate the final result.}}
		\label{fig:parsing}
		\vspace{-4mm}
	\end{center}
\end{figure*}

where $q(G^{(t+1)\rightarrow G^{(t)}})$ and $q(G^{(t)\rightarrow G^{(t+1)}})$ denote the graph state transition probability from one graph to another, and $P(G^{(t+1)}|I; \mathbf{W}, \mathbf{U})$ and $P(G^{(t)}|I; \mathbf{W}, \mathbf{U})$ denote the posterior probability of the graph structure $G^{(t+1)}$ and $G^{(t)}$, respectively. Typically, $P(G^{(t)}|I; \mathbf{W}, \mathbf{U})$ is assumed to follow a Gibbs distribution $\frac{1}{Z}\exp(-\mathcal{L}(F(I, G^{(t)}, \mathbf{W}, \mathbf{U}), Y))$, 	where $Z$ is the partition function, $F(I, G^{(t)}, \mathbf{W}, \mathbf{U})$ is the network prediction, $Y$ is the task-specific target and $\mathcal{L}(\cdot)$ is the corresponding loss function. For example, $Y$ can be the segmentation groundtruth and $\mathcal{L}(\cdot)$ is the pixel-wise cross-entropy loss for the image parsing task. The model is more likely to accept a new graph structure $G^{(t+1)}$ that can bring more significant performance improvement indicated by $\frac{P(G^{(t+1)}|I; \mathbf{W}, \mathbf{U})}{P(G^{(t)}|I; \mathbf{W}, \mathbf{U})}$. The graph state transition probability ratio is computed by:
	
\begin{equation}
\begin{split}
\frac{q(G^{(t+1)\rightarrow G^{(t)}})}{q(G^{(t)\rightarrow G^{(t+1)}})} \propto &\frac{\prod_{<i,j>\in \mathcal{E}^{(t+1)}}(1-(1-p^t_{ij}))}{\prod_{<i,j>\in \mathcal{E}^{(t)}}(1-(1-p^t_{ij}))}\\
&= \prod_{<i,j>\in \mathcal{E}^{(t)}\backslash\mathcal{E}^{(t+1)}}p^t_{ij}.
\label{eq:acceptance}
\end{split}
\end{equation}

The state transition probability is thus calculated by multiplying all merging probabilities of eliminated edges in $G^{(t)}$. It implies that the graph nodes with larger merging probabilities $\{p^t_{ij}\}$ of $G^{(t)}$ are more encouraged to be merged in $G^{(t+1)}$. During testing, the acceptance rate is only determined by the graph state transition probability in Eqn.~\ref{eq:acceptance}. To enable finish the graph structure exploration within a specified time schedule in each step, we can empirically set the upper bound for the sampling trials, say 50 in our experiments.

In the $(t+1)$-th structure-evolving LSTM layer, the information propagation is performed on all nodes with a stochastic node updating sequence along the new graph topology $G^{(t+1)} =< V^{(t+1)}, \mathcal{E}^{(t+1)} > $. The input states $f^{t+1}_i$ for each node $v^{t+1}_i \in V^{t+1}$ are produced by averaging those of all corresponding
merged nodes in $G^{(t)}$. Similarly, the hidden and memory states of $v^{t+1}_i$ are averaged and used for further updating. The weight matrices of the structure-evolving LSTM units are shared for all stacked layers with generated hierarchical graph representations, which helps improve the capability of the network parameters in sensing multi-level semantic abstractions. The final loss for training structure-evolving LSTM includes the final task-related prediction loss and the loss on the predicted merging probabilities for all layers.

\section{Experiments}

The proposed structure-evolving LSTM aims to provide a principled framework to dynamically learn the hierarchal data structures, which is applicable for kinds of tasks (e.g., nature language understanding and image content understanding). However, among all these applications, the semantic object parsing task that requires to produce the pixel-wise labeling by considering the complex interactions between different pixels, superpixels or parts, is a perfect match to better evaluate the structure generation capability of our structure-evolving LSTM. Our dynamically evolved hierarchical graph structures can effectively capture the multi-level and diverse contextual dependencies. We thus evaluate the effectiveness of the proposed structure-evolving LSTM model on the semantic object parsing task (i.e., segmenting an object in the image into its semantic parts) where exploiting multi-level graph representations for the image content is very natural and useful for the final parsing result.

\begin{table*}[!tp]\setlength{\tabcolsep}{2pt}
	\centering
	\caption{Comparison of semantic object parsing performance with several state-of-the-art methods on the PASCAL-Person-Part dataset~\cite{chen2014detect} and with other variants of the structure-evolving LSTM model, including using different LSTM structures, the extracted multi-scale superpixel maps and a deterministic policy with different thresholds for the graph transition, respectively. }\label{tab:person}
	\begin{tabular}{cccccccccccccccccccccc}
		\toprule
		{Method} &  head   &  torso  &  u-arms  & l-arms & u-legs & l-legs & Bkg & Avg \\
		\midrule
		DeepLab-LargeFOV~\cite{chen2014semantic}  & 78.09 & 54.02 & 37.29 & 36.85 & 33.73 & 29.61 & 92.85 & 51.78 \\
		DeepLab-LargeFOV-CRF~\cite{chen2014semantic}  & 80.13 & 55.56 & 36.43 & 38.72 & 35.50 & 30.82 & 93.52 & 52.95 \\
		HAZN~\cite{xia2015zoom} & {80.79} & {59.11} & {43.05} & {42.76} & 38.99 & 34.46 & 93.59 & 56.11\\
		Attention~\cite{chen2015attention} & {-} & {-} & {-} & {-} & - & - & - & 56.39\\
		\midrule
		Grid LSTM~\cite{gridlstm} & 81.85 & 58.85 & 43.10 & 46.87 & 40.07 & 34.59 & 85.97 & 55.90\\
		Row LSTM~\cite{van2016pixel} & 82.60 & 60.13 & 44.29 & 47.22 & 40.83 & 35.51 &　87.07 & 56.80\\
		Diagonal BiLSTM~\cite{van2016pixel} & 82.67 & 60.64 & 45.02 & 47.59 & 41.95 & 37.32 & 88.16 & 57.62\\
		LG-LSTM~\cite{liang2015semantic} & {82.72} & 60.99 & 45.40 & {47.76} & 42.33 & 37.96 & 88.63 & 57.97\\ 
		{Graph LSTM}~\cite{liang2016semantic} & {82.69} & {62.68} & {46.88} & {47.71} & {45.66} & {40.93} & {94.59} & {60.16} \\
		\midrule
		Graph LSTM (multi-scale superpixel maps)~\cite{liang2016semantic} & 83.93 & 64.67 & 48.79 & \textbf{49.44} & 46.57 & 41.38 & 92.36 & 61.02\\
		\midrule
		Structure-evolving LSTM (deterministic 0.5) & 82.93 & 62.59 & 46.91 & 48.06 & 44.73 & 40.39 & 91.77 & 59.63\\
		Structure-evolving LSTM (deterministic 0.7) & \textbf{84.16} & 66.16 & 49.90 & 48.24 & 48.29 & 44.13 &  94.53 & 62.20\\
		Structure-evolving LSTM (deterministic 0.9) & 83.52 & 64.17 & 48.39 & 49.02 & 46.26 & 42.20 & 93.36 & 60.99\\
		\midrule
		\textbf{Structure-evolving LSTM} & 82.89 & \textbf{67.15} & \textbf{51.42} & 48.72 & \textbf{51.72} & \textbf{45.91} & \textbf{97.18} & \textbf{63.57}\\
		
		\bottomrule
	\end{tabular}%
	\vspace{-4mm}
\end{table*}%

\subsection{Semantic Object Parsing Task}

We take the object parsing task as our application scenario, which aims to generate pixel-wise semantic part segmentation for each image, as shown in Fig.~\ref{fig:parsing}. The initial graph $G^{(0)}$ is constructed on superpixels that are obtained through image over-segmentation using SLIC~\cite{achanta2010slic} following~\cite{liang2016semantic}. Each superpixel indicates one graph node and each graph edge connects two spatially neighboring superpixel nodes. The input image first passes through a stack of convolutional layers to generatt convolutional feature maps. The input features $f^0_i$  of each graph node $v_i$ are computed by averaging the convolutional features of all the pixels belonging to the same superpixel node $v_i$. Five structure-evolving LSTM layers are then stacked to learn multi-level graph representations by stochastically grouping some nodes into a large node with the coherent semantic meanings through a bottom-up process. 

To make sure that the number of the input states for the first LSTM layer is compatible with that of the following layers, the dimensions of hidden and memory states in all LSTM layers are set the same as the feature dimension of the last convolutional layer before the LSTM stack. After that, one prediction layer with several $1\times1$ convolution filters produces confidence maps for all labels. During training, we use the groundtruth semantic edge map defined over all the superpixels to supervise the prediction of merging probabilities of all the edges in each LSTM layer. Specifically, the ground-truth merging probability of two graph nodes is set as 1 only if they belong to the same semantic label. L2-norm loss is employed for the back-propagation. The cross-entropy loss is employed on all the predictions layers to produce the final parsing result.

\subsection{Datasets and Implementation Details}

\textbf{Dataset:} We validate the effectiveness of the structure-evolving LSTM on three challenging image parsing datasets. The PASCAL-Person-part dataset~\cite{chen2014detect} concentrates on the human part segmentation on images from PASCAL VOC 2010. Its semantic labels consist of Head, Torso, Upper/Lower Arms, Upper/Lower Legs, and one background class. 1,716 images are used for training and 1,817 for testing. The Horse-Cow parsing dataset is a part segmentation benchmark introduced in~\cite{wang2014semantic}. It includes 294 training images and 227 testing images and each pixel is labeled as head, leg, tail or body. The third task, human parsing aims to predict every pixel with 18 labels: face, sunglass, hat, scarf, hair, upper-clothes, left-arm, right-arm, belt, pants, left-leg, right-leg, skirt, left-shoe, right-shoe, bag, dress and null. Originally, 7,700 images are included in the ATR dataset~\cite{ATR}, with 6,000 for training, 1,000 for testing and 700 for validation. 10,000 images are further collected by~\cite{Co-CNN} to cover images with more challenging poses and clothes variations.

\textbf{Evaluation metric:} The standard intersection over union (IOU) criterion and pixel-wise accuracy are adopted for evaluation on PASCAL-Person-Part dataset and Horse-Cow parsing dataset, following~\cite{chen2014detect}. We use the same evaluation metrics as in~\cite{ATR,Co-CNN} for evaluation on the human parsing dataset, including accuracy, average precision, average recall, and average F-1 score.

\textbf{Network architecture:} For fair comparison with~\cite{chen2014semantic,xia2015zoom,chen2015attention}, our network is based on the publicly available model, “DeepLab-CRF-LargeFOV"~\cite{chen2014semantic} for the PASCAL-Person-Part and Horse-Cow parsing dataset, which slightly modifies VGG-16 net~\cite{simonyan2014very} to FCN~\cite{long2014fully}. “Co-CNN" structure~\cite{Co-CNN} is used to compare with~\cite{ATR,Co-CNN} on one human parsing datasets for fair comparison.

\begin{table*}[!tp]\setlength{\tabcolsep}{3pt}
	\centering
	\caption{Performance comparison with using different numbers of structure-evolving LSTM layers. }\label{tab:step}
	\begin{tabular}{c|c|c|c|c|ccccccccccccccccc}
		\toprule
		{Settings} &  1-layer   &  2-layer  &  3-layer  & 4-layer & Structure-evolving LSTM (full)\\
		\midrule
		Average IoU & 58.19 & 60.23 & 62.59 & 63.18 & \textbf{63.57}\\
		\bottomrule
	\end{tabular}%
\end{table*}%

\begin{table*}[!tp]\setlength{\tabcolsep}{3pt}
	\centering
	\caption{Performance comparison for the predictions by using different levels of graph structures. }\label{tab:level}
	\begin{tabular}{c|c|c|c|c|c|cccccccccccccccc}
		\toprule
		{Settings} &  1st level   &  2nd level  &  3rd level  & 4th level & 5th level & Structure-evolving LSTM (full)\\
		\midrule
		Average IoU & 57.19 & 61.29 & 60.13 & 59.87 & 59.23 & \textbf{63.57}\\
		\bottomrule
	\end{tabular}%
\end{table*}%

\begin{table}[!tp]\setlength{\tabcolsep}{2.8pt}
	\centering\scriptsize
	\caption{Comparison of object parsing performance with five state-of-the-art methods over the Horse-Cow object parsing dataset~\cite{wang2014semantic}. }\label{tab:horsecow}
	\begin{tabular}{cccccccccccccccccccccc}
		\toprule
		& & & & \textbf{Horse} & & &\\
		\hline
		{Method} &  Bkg   &  head  &  body  & leg & tail & Fg & IOU & Pix.Acc \\
		\midrule
		SPS~\cite{wang2014semantic}  & 79.14 & 47.64 & 69.74 & 38.85 & - & 68.63 & - & 81.45 \\
		HC~\cite{hariharan2014hypercolumns}  & 85.71 & 57.30 & 77.88 & 51.93 & 37.10 & 78.84 & 61.98 & 87.18 \\
		Joint~\cite{wang2015joint} & 87.34 & 60.02 & 77.52 & 58.35 & {51.88} & 80.70 & 65.02 & 88.49\\
		{LG-LSTM}~\cite{liang2015semantic} & {89.64} & {66.89} & {84.20} & {60.88} & 42.06 & {82.50} & {68.73} & {90.92} \\
		HAZN~\cite{xia2015zoom} & {90.87} & {70.73} & {84.45} & {63.59} & 51.16 & {-} & {72.16} & {-} \\
		{Graph LSTM}~\cite{liang2016semantic} & {91.73} & {72.89} & {86.34} & {69.04} & {53.76} & {87.51} & {74.75} & {92.76} \\
		\midrule
		\textbf{Ours} & \textbf{92.51} & \textbf{74.89} & \textbf{87.55} & \textbf{71.93} & \textbf{57.45} & \textbf{88.76} & \textbf{76.87} & \textbf{93.45} \\
		\midrule
		& & & & \textbf{Cow} & & &\\
		\hline
		{Method} &  Bkg   &  head  &  body  & leg & tail & Fg & IOU & Pix.Acc \\
		\midrule
		SPS~\cite{wang2014semantic}  & 78.00 & 40.55 & 61.65 & 36.32 & - & 71.98 & - & 78.97 \\
		HC~\cite{hariharan2014hypercolumns}  & 81.86 & 55.18 & 72.75 & 42.03 & 11.04 & 77.04 & 52.57 & 84.43 \\
		Joint~\cite{wang2015joint} & 85.68 & 58.04 & 76.04 & 51.12 & 15.00 & 82.63 & 57.18 & 87.00\\
		{LG-LSTM}~\cite{liang2015semantic} & {89.71} & {68.43} & {82.47} & {53.93} & {19.41} & {85.41} & {62.79} & {90.43}\\
		HAZN~\cite{xia2015zoom} & {90.66} & \textbf{75.10} & {83.30} & {57.17} & 28.46 & {-} & {66.94} & {-} \\
		{Graph LSTM}~\cite{liang2016semantic} & {91.54} & {73.88} & {85.92} & \textbf{63.67} & {35.22} & {88.42} & {70.05} & {92.43} \\
		\midrule
		\textbf{Ours} & \textbf{92.88} & \textbf{77.75} & \textbf{87.91} & \textbf{67.60} & \textbf{42.86} & \textbf{90.71} & \textbf{73.80} & \textbf{93.57} \\
		\hline
				\vspace{-2mm}
	\end{tabular}%
\end{table}%

\textbf{Training:} The SLIC over-segmentation method~\cite{achanta2010slic} generates 1,000 superpixels on average for each image. The learning rate of the newly added layers over pre-trained models is initialized as 0.001 and that of other previously learned layers is initialized as 0.0001. All weight matrices used in the structure-evolving LSTM units are randomly initialized from a uniform distribution of [-0.1, 0.1]. We only use five LSTM layers for all models since only slight improvements are observed by using more LSTM layers, which also consumes more computation resources. The weights of all convolutional layers are initialized with Gaussian distribution with standard deviation 0.001. We train all the models using stochastic gradient descent with a batch size of 1 image, momentum of 0.9, and weight decay of 0.0005. We fine-tune the networks on “DeepLab-CRF-LargeFOV" and train the networks based on “Co-CNN" from scratch for roughly 60 epochs. The structure-evolving LSTM is implemented by extending the Caffe framework~\cite{jia2014caffe}. All networks are trained on a single NVIDIA GeForce GTX TITAN X GPU with 12GB memory. In the testing stage, extracting superpixels takes 0.5s and our method takes 1.3s per image in total.   

\subsection{Results and Comparisons}

\textbf{Comparisons with State-of-the-art Methods.} We report the result comparisons with recent state-of-the-art methods on PASCAL-Person-part dataset, Horse-Cow parsing dataset and ATR dataset in Table~\ref{tab:person}, Table~\ref{tab:horsecow}, Table~\ref{tab:atr}, respectively. The proposed structure-evolving LSTM structure substantially outperforms these baselines in terms of most of the metrics, especially for small semantic parts. This superior performance achieved by the structure-evolving LSTM demonstrates the effectiveness of capturing multi-scale context by propagating information on the generated graph structures.

\textbf{Comparisons with Existing LSTM Structures.} Table~\ref{tab:person} gives the performance comparison among different LSTM structures, including Row LSTM~\cite{van2016pixel}, Diagonal BiLSTM~\cite{van2016pixel}, LG-LSTM~\cite{liang2015semantic}, Grid LSTM~\cite{gridlstm} and Graph LSTM~\cite{liang2016semantic}, which use the same network architecture and number of LSTM layers. In particular, Row LSTM, Diagonal BiLSTM, LG-LSTM, Grid LSTM and LG-LSTM use the fixed locally factorized topology for all images while Graph LSTM propagates information on the fixed superpixel graph. It can be seen that exploiting the multi-level graph representations for different LSTM layers leads to over 3.41\% improvement than the pre-defined LSTM structures on average IoU.

\begin{table}[!tp]\setlength{\tabcolsep}{2.8pt}
	\centering\scriptsize
	\caption{Performance comparison with state-of-the-art methods when evaluating on ATR dataset~\cite{ATR}. Following~\cite{Co-CNN}, we also take the additional 10,000 images in ~\cite{Co-CNN} as extra training images, denoted as ``Ours (more data)". Comparison of human parsing performance with seven state-of-the-art methods when evaluating on ATR dataset.}\label{tab:atr}
	\begin{tabular}{cccccccccccccccccccccc}
		\toprule
		\textbf{Method} &  \textbf{Acc.}   &  \textbf{F.g. acc.}  &  \textbf{Avg. prec.}   &    \textbf{Avg. recall}  &  \textbf{Avg. F-1 score} \\
		\midrule
		Yamaguchi et al.~\cite{yamaguchi2012parsing}& 84.38 & 55.59 & 37.54 & 51.05 & 41.80 \\
		PaperDoll~\cite{Yamaguchiparsing13} & 88.96 & 62.18 & 52.75 & 49.43 & 44.76 \\
		{M-CNN}~\cite{M-CNN} &{89.57} &{73.98} &{64.56} &{65.17} &{62.81}\\
		ATR~\cite{ATR} & {91.11} & {71.04} & {71.69} & {60.25} & {64.38}\\
		{Co-CNN}~\cite{Co-CNN} & 95.23 & 80.90 & 81.55 & 74.42 & 76.95\\
		{Co-CNN (more)}~\cite{Co-CNN} & {96.02} & {83.57} & {84.95} & {77.66} & {80.14}\\
		{LG-LSTM}~\cite{liang2015semantic} & {96.18} & {84.79} & {84.64} & {79.43} & {80.97}\\
		{LG-LSTM (more)}~\cite{liang2015semantic} & {96.85} & {87.35} & {85.94} & {82.79} & {84.12}\\
		CRFasRNN (more)~\cite{crfasrnn} & {96.34} & {85.10} & {84.00} & {80.70} & {82.08}\\
		{Graph LSTM} & {97.60} & {91.42} & {84.74} & {83.28} & {83.76}\\
		{Graph LSTM (more)} & {97.99} & {93.06} & {88.81} & {87.80} & {88.20}\\
		\midrule
		{Ours} & {97.71} & {91.76} & {89.37} & {86.84} & {87.88}\\
		\textbf{Ours (more)} & \textbf{98.30} & \textbf{95.12} & \textbf{90.08} & \textbf{91.97 } & \textbf{90.85}\\
		\bottomrule
		\vspace{-4mm}
	\end{tabular}%
\end{table}%

\begin{figure*}[!tp]
	\begin{center}
		\includegraphics[scale=0.53]{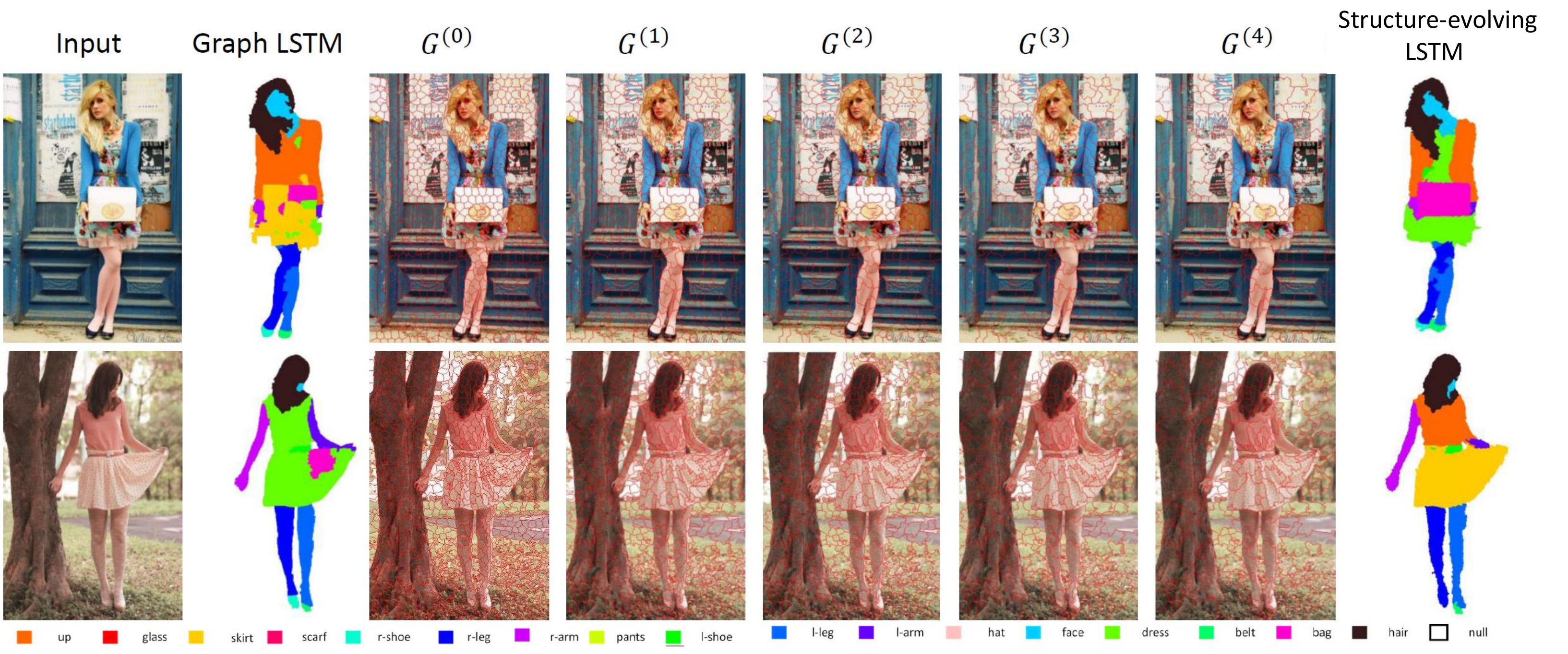}
		\caption{{Comparison of parsing results of our structure-evolving LSTM and Graph LSTM on ATR dataset and the visualization of the corresponding generated multi-level graph structures. Better viewed in zoomed-in color pdf.}}
		\label{fig:results}
		\vspace{-6mm}
	\end{center}
\end{figure*}

\textbf{Discussion on Using Stochastic Policy.} Note that the structure-evolving LSTM stochastically merges some graph nodes and employs an acceptance rate to determine whether a new graph structure should be accepted. An alternative way is deterministically merging some graph nodes by hard-thresholding, that is, two nodes are merged only if their merging probability is larger than a fixed threshold $T$. In our experiment, three thresholds (i.e., 0.5, 0.7,0.9) are tested in Table~\ref{tab:person}. Using a smaller threshold (e.g., 0.5) is more likely to obtain more aggressive graph transitions by merging more nodes while a larger threshold would prevent the graph from changing its structure. It is shown that using 0.7 threshold in the deterministic policy obtains the best performance, which is still inferior to the proposed stochastic policy. Additionally, we find that only slight performance differences are obtained after running the feed-forward prediction using the structure-evolving LSTM for ten times, which verifies the robustness of the structure-evolving LSTM.

\textbf{Comparisons with Using All Pre-defined Graph Structures.} An optional strategy to capture multi-scale context is to utilize pre-computed multi-scale superpixel maps as the intermediate graph structures, reported as “Graph LSTM (multi-scale superpixel maps)" in Table~\ref{tab:person}. Five predefined graph structures in LSTM layers can be constructed by five superpixel maps with 1000, 800, 600, 256 400, 200 superpixels, respectively. These superpixel numbers are consistent with the averaged node number of our learned graph structures for all training images. The superiority of ``Structure-evolving LSTM" demonstrates that exploiting adaptive graph structures makes the structure more consistent with the high-level semantic representation instead of just relying on the bottom-up oversegmentation.

\textbf{Discussion on Predictions with Different Levels of Graphs.} The performance of using different numbers of the structure-evolving LSTM layers is reported in Table~\ref{tab:step}. It demonstrates that exploiting more levels of graph structures makes the network parameters learn different levels of semantic abstraction, leading to better parsing results, whereas the previous LSTM model~\cite{liang2016semantic} reported that no performance gain is achieved with more than two LSTM layers. Note that the parsing prediction is produced by each LSTM layer and these predictions are element-wisely summed to generate the final result. The individual parsing performance by using each graph structure is reported in Table~\ref{tab:level}. The higher-level graph structure may wrongly merge bottom-up graph nodes, which thus may lead to the deteriorated performance. However, combining all predictions from all the structure-evolving LSTM layers can largely boost the prediction benefited from incorporating the multi-scale semantical context.

\textbf{Visualization.} The qualitative comparisons of parsing results on ATR dataset and the graph structures exploited by structure-evolving LSTM layers are visualized in Fig.~\ref{fig:results}. The structure-evolving LSTM outputs more reasonable results for confusing labels (e.g., skirt and dress) by effectively exploiting multi-scale context with the generated multi-level graph structures.

\section{Conclusion}

We presented a novel interpretable structure-evolving Graph LSTM which simultaneously learns multi-level graph representations for the data and LSTM network parameters in an end-to-end way. While following the line of graph-based RNNs, our work significantly improves the way of network learning by allowing the underlying multi-level graph structures to evolve along with the parameter learning. The network can thus learn representations to better fit the hidden structure of the data. Moreover, we propose a principled approach to evolve graph structures stochastically, which is not straightforward and could have potential impact on the application of graph-based RNNs in multiple domains. We have demonstrated its effectiveness on the object parsing task for an image. In future, the structure-evolving LSTM can be extended to enable the reversible graph transition (e.g., splitting some merged nodes) during the LSTM network optimization. We will also evaluate its performance on the tasks of other modalities, such as the social networks.
{\small
\bibliographystyle{ieee}
\bibliography{egbib}
}

\end{document}